\documentclass[]{spie}  

\usepackage{amsmath,amsfonts,amssymb}
\usepackage{graphicx}
\usepackage[colorlinks=true, allcolors=blue]{hyperref}
\usepackage{comment}
\usepackage{float} 
\usepackage{subfig}

\title{Distributed Transmission Control for Wireless Networks using Multi-Agent Reinforcement Learning}

\author[a]{Collin Farquhar}
\author[a]{Prem Sagar Pattanshetty Vasanth Kumar}
\author[a]{Anu Jagannath}
\author[a]{Jithin Jagannath}

\affil[a]{Marconi-Rosenblatt AI/ML Innovation Lab, ANDRO Computational Solutions,\newline 7980 Turin Rd, Beeches Technical Campus Building 1, Rome, NY, USA}

\authorinfo{Further author information: (Send correspondence to C.F.)\\C.F.: E-mail: farquhar13@gmail.com\\J.J.: E-mail: jjagannath@androcs.com}

\begin{document} 
\maketitle

\begin{abstract}
We examine the problem of transmission control, i.e., when to transmit, in distributed wireless communications networks through the lens of multi-agent reinforcement learning. Most other works using reinforcement learning to control or schedule transmissions use some centralized control mechanism, whereas our approach is fully distributed. Each transmitter node is an independent reinforcement learning agent and does not have direct knowledge of the actions taken by other agents. We consider the case where only a subset of agents can successfully transmit at a time, so each agent must learn to act cooperatively with other agents. An agent may decide to transmit a certain number of steps into the future, but this decision is not communicated to the other agents, so it the task of the individual agents to attempt to transmit at appropriate times. We achieve this collaborative behavior through studying the effects of different actions spaces. We are agnostic to the physical layer, which makes our approach applicable to many types of networks. We submit that approaches similar to ours may be useful in other domains that use multi-agent reinforcement learning with independent agents.
\end{abstract}


\section{INTRODUCTION}
\label{sec:intro}  

Modern communication environments can be large, complex, and dynamic, and thus can pose a problem for traditional algorithms that are often designed for a particular goal or environment, and may have difficulty scaling and adapting to new circumstances. Additionally, many traditional approaches rely on a centralized controller, and without such a controller may fail to find optimal performance if a distributed approach is employed. Machine learning and reinforcement learning approaches have been proposed as a promising alternative as they may learn to find solutions even in for extremely challenging complex and dynamic environments \cite{Jagannath_2019}. Herein, we present a Multi-Agent Reinforcement Learning (MARL) approach applied to the problem of distributed transmissions control. 

Our distributed approach uses individual agents to control the transmission of devices (if and when to transmit), providing a solution to transmission control and medium access control tasks where a centralized control mechanism is infeasible or disadvantageous. Distributed approaches have the advantage of limiting communication overhead and may be more extensible in terms of the number of devices and spacial range, as a centralized bottleneck is avoided and devices do not need to be near a centralized controller. Additionally, in some systems, especially large or sensitive systems, a centralized approach may not be feasible in terms of security and robustness. For example, an attack on a centralized mechanism may cause the entire network to fail. Communication settings where this distributed approach may be advantageous include tactical ad hoc networks, sensor networks, device-to-device communications, and various other internet of things systems.

Moving away from centralized approaches poses some technical challenges. In general, network-level planning is easier with a centralized mechanism. From a reinforcement learning (RL) perspective, if centralized mechanism controls each node's actions and has complete access to all the nodes' observations and rewards, then the problem reduces to a classical Markov Decision Process (MDP) and single-agent solutions can be used \cite{zhang2018fully}. 

We consider the setting with $n$ transmitters represented as individual reinforcement learning agents. The environment is such that only a limited number of the agents are able to successfully transmit concurrently due to the friendly interference caused by transmitting. It is then the task of the agents to learn when to transmit. We devise a general form for the action space and analyze the effects of different actions spaces sizes on the performance in terms of network throughput and fairness. By tailoring the action space to the environment we see that the agents are able to learn cooperative transmission patterns that lead to efficient and fair performance. Based on our analysis, we choose action spaces that perform well and compare them against CSMA-style benchmarks. We find that our RL agents often outperform the benchmarks in terms of throughput and fairness.

We test our approach in a custom simulation environment based on OpenAI Gym \cite{Gym}. Our environment is open source and is freely available for download on GitHub with a link provided in Appendix \ref{Link to Software}. 

\section{Related Work}

A similar objective to ours is explored in  Ref.~\citenum{jiang2019learning} from the perspective of domain-agnostic cooperative RL. The work proposes a method for optimizing the trade-off between fairness and efficiency in MARL, which is analogous to our desire to optimize optimize fairness and throughput. However, the agents are allowed to share information with other local agents. In the context of communications systems, however, this information sharing may induce unwanted overhead, and in our approach we do not rely on information sharing.

An adjacent area or research involves \textit{learning to communicate} for RL agents --- that is, learning how to share information so that agents can cooperatively achieve some goal. \cite{foerster2016learning, kim2019learning}.  While this goal may not be related to communications systems per se, there is an inherent scheduling component for the information sharing which is similar to our task. In particular,  Kim et al. \cite{kim2019learning} considers an environment with bandwidth and medium access constraints. While these works use distributed execution, they rely on a centralized training mechanism. In contrast, in our work we focus on optimizing communications systems using fully distributed MARL. 

Ref.~\citenum{Hetero_MAC} investigates how a deep reinforcement learning agent can learn a Medium Access Control (MAC) strategy to coexist with other networks that use traditional (not machine learning--based) MAC algorithms. In some ways, it may be easier to learn to coexist with a traditional MAC algorithms, as the algorithms may have a certain structures, regularities, or periodicities that can be learned by the agent. In this paper we focus purely on multi-agent scheduling, where all agents have the same action spaces and it is difficult to predict a priori what patterns may emerge.

Despite not using a machine learning approach, perhaps the closest to this paper is the recent work of Ref. \citenum{Distributed_Fair_Scheduling} which proposes an algorithm for distributed, fair MAC scheduling and additionally provides theoretical guarantees for service disparity.

\section{Problem Formulation and Approach}

We consider scenarios with $n$ agents acting as transmitting devices (nodes in a wireless network). The transmission medium is limited such that only a threshold number, $k$, of agents can successfully transmit at the same time. The threshold is uniform for all agents: if more than $k$ agents transmit on the same step, then all of the transmissions are unsuccessful at the respective receivers. Given this threshold behavior, we are agnostic to the PHY-layer. With $k>1$ we can model an abstracted CDMA PHY-layer where $k$ can be interpreted as the spreading factor. For $k=1$ we can model many other systems that use an OFDM-based PHY-layer.

The agents may only transmit once on a given step. We leave the definition of a transmission abstracted so that a transmission could represent either the transmission of a single packet, multiple packers, or some number of bits.

\begin{figure} 
\begin{center}

\includegraphics[height=6cm]{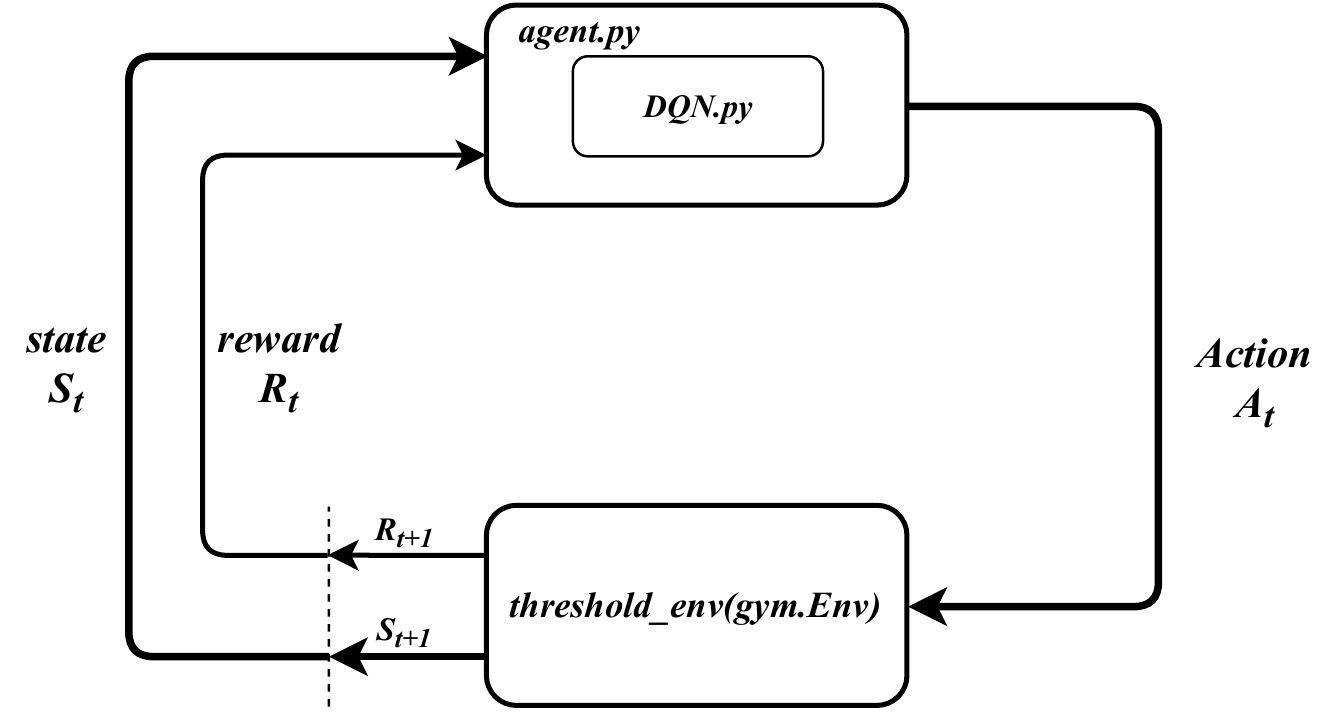}

\end{center}
\caption[RlFlow] { \label{fig:RlFlow} Depiction of the flow of information between our software during the reinforcement learning process. The \textit{threshold\_env} inherits from \textit{gym.Env}, the base class for OpenAI Gym, and serves as the environment in which the agents act. In \textit{agent.py} the agents are defined using a DQN model. State, reward, and action information ($S_t$, $R_t$, and $A_t$) are passed between the environment and the agents. Since this is a multi-agent setting, $S_t$, $R_t$, and $A_t$ are lists containing the respective information for each agent. E.g., $S_t = [s_{t}^{(1)}, s_{t}^{(2)}, s_{t}^{(3)}, ...]$ is a list of the states for each agent at time $t$. Figure adapted from Ref. \citenum{sutton2018reinforcement}.}
\end{figure}

\subsection{Simulation Software}
We provide a software platform to model the interaction of agents in a threshold-limited communication setting. A link to view or download the code can be found in Appendix \ref{Link to Software}. There are two primary components to the software, the first is \textit{agent.py} where the agents' RL models are defined and used to select actions, and the second is the simulation environment, called \textit{threshold\_env}, which is built on top of OpenAI Gym and inherits from the \textit{gym.Env} base class. The primary functionality is implemented in the \textit{step} method, which advances the simulation. The \textit{step} method takes a list of actions (one action per agent) as input. The actions are performed in the environment and the agents move into a new state and receive some reward. The information regarding the new states and the reward is then returned by the \textit{step} method to be used by the agents' RL models in the \textit{agent.py} file. The information flow between our software is depicted in Figure \ref{fig:RlFlow}.

While we use our custom environment in this paper as it is suits our desired level of abstraction, our approach can easily be tested using other dedicated wireless communication simulators if one wanted to more finely model the particularities a specific communication system. Internally, we have tested our approach with ns3-gym \cite{ns3gym} and found the results to be consistent with our custom environment.

\subsection{RL Background}
A general reinforcement learning problem is formulated as a Markov Decision Process (MDP), which may be represented by the tuple $(\mathbb{S}, \mathbb{A}, P, R)$ where $\mathbb{S}$ is the set of states an agent may encounter, $\mathbb{A}$ is the set of actions an agent can take, $P(s_{t + 1} | s_t, a_t)$ is the state transition function which follows the Markov Property and gives the probability that the agent's next state is $s_{t + 1}$ given that the agent is in state $s_t$ and takes action $a_t$, and $R$ is the reward function which outputs a real number. It is then the task of an agent to maximize it's long-term reward over many steps.

Our environment can be classified as a multi-agent, Partially Observable Markov Decision Process (POMDP). Each agent is taking actions regarding if and when to transmit. The actions taken by an agent are not directly communicated to the other agents in the environment. However, the other agents can indirectly observe the effects of the these actions by measuring the overall level of interference. In the following sections, we define the state space, action space, reward function, and the RL model.

\subsection{State Space}
An agent's observation of their state consists of four features (t, s, i, b) which are described below. 

\begin{itemize}
    \item Whether the agent transmitted, $t \in \{0, 1\}$
    \item Whether the transmission was successful, $ s \in \{0, 1\}$
    \item Interference-sensed (normalized), $i \in [0, 1]$
    \item buffer size (normalized), $b \in [0, 1]$
\end{itemize}

If an agent did not transmit then $s$ is automatically set to $0$. The interference-sensed feature is normalized by dividing the number of transmissions from other agents by the number of other agents. We constrain the interference-sensed feature, $i$, such that an agent cannot transmit and sense on the same step, as in a realistic system its transmission would dominate any local sensing mechanism. If an agent transmits, then $i$ is set to $0$. Buffer size can be thought of as representing how many packets are waiting to be transmitted by an agent. The buffer size feature, $b$, is normalized by setting a maximum buffer size.

\subsection{Action Space} \label{Define Action Space}
We provide a \textit{structure} for the action space. In this work, we test actions spaces of different sizes, but they all follow the same form. Effectively, the agent has the choices: do not transmit on the next step, transmit on the next step, or wait for certain number of steps and then transmit. The encoding for a selected action $a_i$ is shown below. 

\begin{itemize}
    \item $a_0$ $\rightarrow$ Do not transmit 
    \item $a_1$ $\rightarrow$ Transmit on the next step
    \item $a_2$ $\rightarrow$ Transmit after one step
    \item $a_3$ $\rightarrow$ Transmit after two steps
    \item $a_4$ $\rightarrow$ Transmit after three steps 
    \item ...
\end{itemize}

A transmission lasts for one simulation-step. If an agent's buffer is zero, then the RL model is not used and the agent takes no action as there are no packets to be transmitted, and thus no decision to be made. Since we consider multi-agent settings, the agents will be making decisions asynchronously as some actions last over multiple steps. In section \ref{Action Space Analysis} we analyze the performance of different action space sizes. An action space of size $m$, ($|\mathbb{A}| = m$), is defined as $\mathbb{A} = \{a_i\}_{i = 0}^{m - 1}$.

\subsection{Reward Function}
The reward for function for single-step actions is shown below. 

\begin{equation}
R = \begin{cases} 
      1 & \text{successful transmission}\\
      -1 & \text{unsuccessful transmission} \\
      0 & \text{no transmission} \\
   \end{cases}
   \label{eqn:reward}
\end{equation}
For multi-step actions, the single-step rewards are averaged over the number of steps the action acts over.

\subsection{RL Model}
We use the vanilla Deep Q-Network (DQN) algorithm as the RL model for each agent. At a high level, the algorithm consists of a neural network that takes an agent's state observation as input and outputs an estimate of the Q-value for each possible action. The Q-value represents the long-term sum of rewards an agent will receive if they take a particular action in a particular state and then from that point on continue to act optimally with respect to the Q-value estimates. 

Unless the agent selects an action randomly according to it's exploration mechanism, the agent will select the action $a_t$ that corresponds to the maximum Q-value. After taking the step in the environment where the agent performs the action, the transition $(s_t, a_t, r_t, s_{t + 1})$ is saved to the agent's experience replay memory (assuming a single-step action). The experience replay memory is sampled during training. If an agent selects an action that extends over multiple steps, the state at the time of the decision, $s_t$ is saved, and the per-step reward is averaged over the duration of the action. For example, if an action taken at time $t$ extends over $n$ steps, then let $\tilde{r} = \frac{1}{n}\sum_{i=t}^{t + n - 1} r_i$, and the transition that is saved is $(s_t, a_t, \tilde{r}, s_{t + n})$.

If an agent's buffer size is zero, then the RL model is not used since there are no packets to send and no decision to make; i.e., the agent takes no action and no transition is saved to the agent's memory.

The DQN hyperparameters are described in Appendix \ref{DQN Hyperparameters} and the code is provided in the repository linked to in Appendix \ref{Link to Software}.

\section{Experiments and Results}

In this section we describe a number of experiments and analyze the agents' performance. We start in Sec. \ref{Action Space Analysis} by investigating the effect of the action space on throughput and fairness for different number of agents and different threshold values for the maximum number of concurrent successful transmissions, $k$. Based on these results, we provide an action space recommendation that we use for the remaining experiments where we compare agent performance versus CSMA-inspired benchmarks in Sec. \ref{Benchmarking}. Lastly, in Sec. \ref{Buffer Experiments}, we examine the how the advantage the RL agents provide relates to an agent's ability to manage its buffer size for different buffer rates (source rates). All experiments are run 3 times for 10,000 steps and the results shown are averages over the 3 runs.

Throughput is calculated as the number of successful transmissions per step. Since an agent may only transmit one packet per step, the maximum possible network throughput is $k$. For greater clarity, the throughput data are smoothed over $100$ steps. The plotted value at time step $t$ is the average over the $100$ values preceding time step $t$. The fairness at time $t$ is calculated using Jain's fairness index as
\begin{equation}
    F_t = \frac{ (\sum_{i=1}^n x_t^{(i)})^2 }{ n  \sum_{i=1}^n (x_t^{(i)})^2}, 
\end{equation}

where $n$ is the number of agents and $x_t^{(i)}$ is the smoothed throughput for agent $i$ at time $t$. A value of 1 represents maximum fairness.

\subsection{Action Space Analysis} \label{Action Space Analysis}
We test a range of action space sizes for $2$, $4$, and $10$ agents. We also vary the threshold, $k$, for additional tests with $4$ and $10$ agents. The buffer rate is such that the buffer is incremented on every step and is uniform for all agents, so the agents always have packets to transmit. 

\begin{table}[h]
    \centering
    \begin{tabular}{ |c||p{3cm}|p{3cm}||p{3cm}|p{3cm}| }
        \hline
        \multicolumn{1}{|c|}{} &        
        \multicolumn{2}{|c|}{Throughput, $k=2$} &
        \multicolumn{2}{|c|}{Fairness, $k=2$}
        \\
        \hline \hline
        No. of Actions & 4 Agents & 10 Agents & 4 Agents & 10 Agents\\
        \hline
        2 Actions & Mean = 1.23383   STDEV = 0.05380 & Mean = 1.13013   STDEV = 0.09343 & Mean = 0.79449   STDEV = 0.01814 & Mean = 0.38081   STDEV = 0.04988 \\
        \hline
        3 Actions & Mean = 1.84297   STDEV = 0.00767 & Mean = 1.05710   STDEV = 0.05481 & Mean = 0.99914   STDEV = 0.00010 & Mean = 0.53656   STDEV = 0.01878 \\
        \hline
        4 Actions & Mean = 1.29135   STDEV = 0.02454 & Mean = 1.00863   STDEV = 0.06923 & Mean = 0.99897   STDEV = 0.00073 & Mean = 0.71488   STDEV = 0.02333 \\
        \hline
        5 Actions & Mean = 1.14423   STDEV = 0.26474 & Mean = 1.04642   STDEV = 0.06590 & Mean = 0.99289   STDEV = 0.00988 & Mean = 0.70399   STDEV = 0.16237 \\
        \hline
        6 Actions & Mean = 0.90587   STDEV = 0.07312 & Mean = 1.05044   STDEV = 0.19130 & Mean = 0.98259   STDEV = 0.01012 & Mean = 0.89317   STDEV = 0.04956 \\
        \hline
        7 Actions & Mean = 0.86968   STDEV = 0.05590 & Mean = 1.21009   STDEV = 0.17026 & Mean = 0.96761   STDEV = 0.02630 & Mean = 0.91798   STDEV = 0.02299 \\
        \hline
        8 Actions & Mean = 0.86249   STDEV = 0.06013 & Mean = 1.10272   STDEV = 0.23757 & Mean = 0.92256   STDEV = 0.00266 & Mean = 0.91769   STDEV = 0.06550 \\
        \hline
        9 Actions & \center -- & Mean = 1.00109   STDEV = 0.07765 & \center -- & Mean = 0.73018   STDEV = 0.33294 \\
        \hline
        10 Actions & \center -- & Mean = 0.91435   STDEV = 0.12382 & \center -- & Mean = 0.92223   STDEV = 0.05944 \\
        \hline
        11 Actions & \center -- & Mean = 0.96161   STDEV = 0.03941 & \center -- & Mean = 0.94661   STDEV = 0.02341 \\
        \hline
    \end{tabular}
    \caption{Average and standard deviation of network throughput and fairness for 4 and 10 agents with $k = 2$. The values are computed from the last 1,000 steps out of a 10,000 step training run, and are averaged over 3 runs.}
    \label{table:1}
\end{table}

Results for throughput and fairness for 4 and 10 agents with $k = 2$ are show in Table \ref{table:1}. For 4 agents throughout is maximized with an action space of size $|\mathbb{A}| = 3$; fairness is above $0.99$ for 3, 4, and 5 actions. For 10 agents the highest throughput value is achieved with $|\mathbb{A}| = 7$. It is interesting to note that the second best throughput is achieved with $|\mathbb{A}| = 2$, but the corresponding fairness is very poor; fairness is near or above $0.9$ for $|\mathbb{A}| \geq 6$, with the exception of 9 actions, but this may have been due to unlucky sampling as the standard deviation is much larger than that for any other action space size. To manage the trade-off between throughput and fairness for 10 agents, a larger actions space shows better results, as this gives good results for fairness and does not (significantly) diminish throughput performance.

The size of the action space clearly has an effect on throughput and fairness. This fact is both obvious and interesting. An action space of size 2 amounts to a decision of \textit{whether to transmit}. Larger action spaces conceptually change the decision to include: \textit{how long to wait before transmitting}. It may not have been predictable a priori that extending the action space in this way would have such an effect as, fundamentally, on each step an agent is either transmitting or not, and each action $a_i$ for $i > 0$ only schedules one transmission (with $a_0$ being a single-step decision not to transmit).

Viewed another way, extending the action space is a way of biasing the set of actions so that the agents have a lower probability of miscoordination. Ref. \citenum{miscoordination} defines \textit{coordination} in the context of MARL as ``the ability of two or more agents to jointly reach a consensus over which actions to perform in an environment", and \textit{miscoordination penalty} is incurred when the joint actions lead to a lower reward. In our setting, we can define miscoordination to mean when more than $k$ agents transmit on the same step, giving a reward of $-1$ for all agents that transmitted.

For example, if $k = 2$ and actions were selected uniformly at random for 10 agents with $|\mathbb{A}| = 2$ ($a = 0$ meaning ``don't transmit" and $a = 1$ meaning "transmit"), then probability of miscoordination (of more than 2 agents selecting $a = 1$ on a given step) is, by the binomial distribution, 
\begin{equation}
    P(\text{miscoordination}) = 0.5^{10} \sum_{i = k + 1 = 3}^{10} \binom{10}{i} \approx 95\%.
    \label{eqd:P_miscoordination}
\end{equation}

With the assumption of randomly selected actions, as the action space increases the probability of miscoordination decreases. More details and a proof of this can be found in Appendix \ref{Miscoordination Penalty Probabilitiess}. However, as the size of the action space, $|\mathbb{A}|$, keeps increasing, eventually a lower probability of miscoordination would also result in diminishing throughput. Of course, our agents do not select actions uniformly at random, but still, this theoretical analysis implies that there is some optimal middle-ground for $|\mathbb{A}|$.

Additional tabulated results are displayed in Appendix \ref{Action Space Analysis Additional Data}, which together with Table \ref{table:1} help inform us as we formulate a heuristic for finding an optimal action space. We make note of some trends emerging from the empirical data. For example, the fairness data for $k = 1$ in Appendix \ref{Action Space Analysis Additional Data} shows a general trend that fairness increases and then may plateau as the number of actions increase. For throughput, we generally observe that as we increase the number of actions throughput increases until a point, and then may decrease. This trend can be seen clearly for for 4 agents with $k = 2$ agents in Table \ref{table:1}; and for 2 agents with $k = 1$, and 10 agents with $k = 5$ both in Appendix \ref{Action Space Analysis Additional Data}; incidentally, it is when $|\mathbb{A}| = 3$ that throughput is maximized in all the aforementioned examples, but we conjecture that the reason for this is not caused by an inherent property of the 3-action space. Rather, for each example, $3 = \frac{n}{k}+ 1$ where $n$ is the number of agents. 

Based our empirical results we recommend the following heuristic for choosing the size of the action space which we choose to promote good performance for both throughput and fairness; we create $\mathbb{A}$ such that
\begin{equation} \label{eqn:heursitic}
    |\mathbb{A}| = \frac{n}{k}+ 1.
\end{equation}

In some cases, like for 10 agents in Table \ref{table:1}, the throughput is high for the smallest action space, but when this occurs the fairness is unacceptably low. The same behavior can be observed for 10 agents with $k = 1$ in the appendix. In these cases as well, using our heuristic we find a much more equitable trade-off between throughput and fairness.

\subsection{Benchmarking} \label{Benchmarking}
Using our heuristic in Equation (\ref{eqn:heursitic}) to set the size of action space, we compare the performance of our RL agents against three distributed CSMA-inspired benchmarks. 

CSMA (Carrier Sense Multiple Access) is a MAC protocol used in Wi-Fi systems; the premise is that a device will sense the spectrum to see if the transmission medium (channel) is clear. If the channel is clear, then the device may transmit. If the transmission is unsuccessful, because, for example, another device also transmitted at the same time and caused interference, then the device randomly chooses a number of steps, $x$, in some predefined interval, $\{z \in \mathbb{Z} : 0 \leq z < X\}$, to wait before attempting to transmit again. We'll refer to $x$ as a \textit{backoff timer} and $X$ as the \textit{backoff--upper-bound}. 

We consider these algorithms reasonable to compare against for benchmarking as our action space may be viewed as doing something similar: giving the agents the ability to choose how long to wait before attempting a transmission.

We call the three algorithms we use for benchmarking \textit{exponential CSMA}, \textit{p-persistent CSMA}, and \textit{p-persistent}. For \textit{exponential CSMA}, we initially set $X = 2$; if the a transmission is unsuccessful, a new $x$ is randomly selected and the new backoff--upper-bound is set to $2X$. When the agent has a successful transmission, the backoff--upper-bound is reset to $X = 2$. In \textit{p-persistent CSMA}, $X$ is a constant and it set to the size of the action space $|\mathbb{A}|$ that the RL agent is using in the benchmarking experiment. The \textit{p-persistent} algorithm works the same as \textit{p-persistent CSMA}, with the only difference being that a spectrum reading that shows no transmissions is not required as a precondition to transmit, as it is for the other two algorithms.

\begin{figure}[t]
    \centering
    \subfloat[Throughput for 10 agents, $k = 5$, $|\mathbb{A}| = 3$.]{
    {\includegraphics[width=8cm]{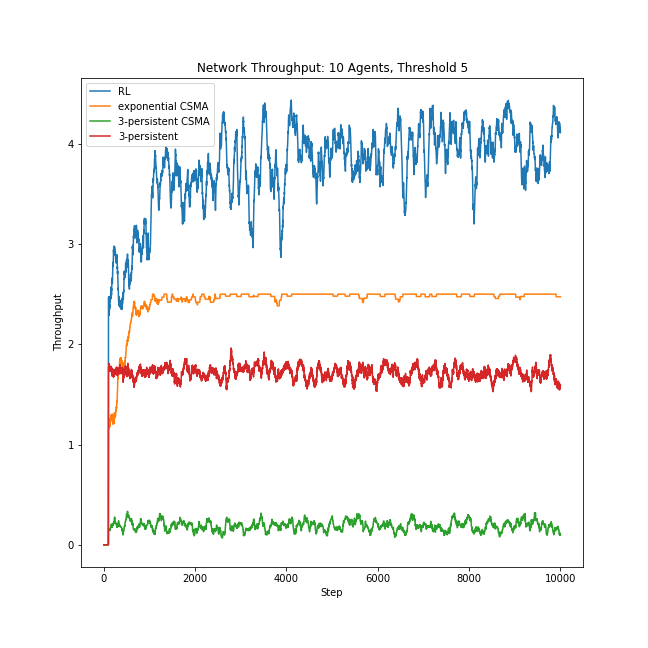} }
    }
    \enspace
    \subfloat[Fairness for 10 agents, $k = 5$, $|\mathbb{A}| = 3$.]{
    {\includegraphics[width=8cm]{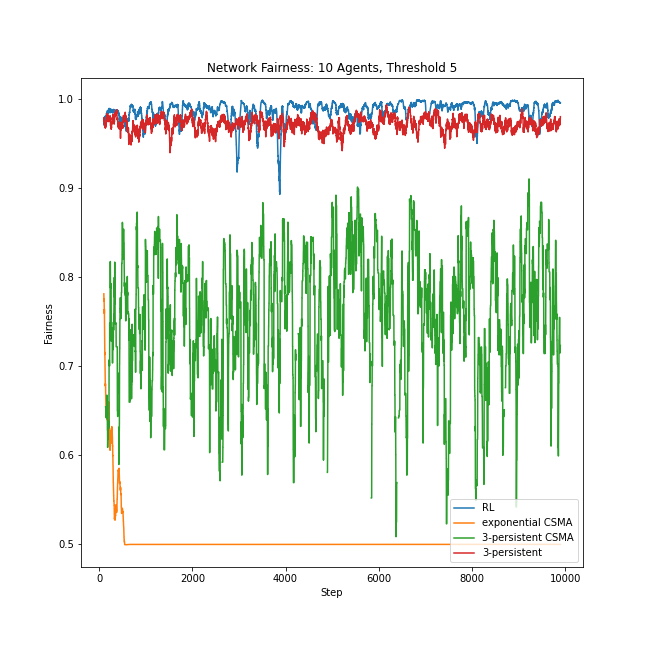} }
    }
    \caption{Network throughput and fairness with the performance for the RL agents and the benchmarking algorithms plotted against each other. The plotted data is the averaged over three runs.}
    \label{fig:benchmarking}
\end{figure}

Figure \ref{fig:benchmarking} shows plots for throughput and fairness for an experiment with 10 agents, $k = 5$, and as in the previous section, the agents' buffers are incremented on every step. Using Equation (\ref{eqn:heursitic}) we set the size of the action space, $|\mathbb{A}| = 3$. We see that our RL agents achieve nearly optimal network performance, as a fairness of 1 is optimal, and a network throughput of $k$ is optimal. The network throughput for the benchmarking algorithms are comparatively low. We see that \textit{3-persistent CSMA} produces the lowest throughput, while the \textit{3-persistent} algorithm does better, which is likely due to the fact that it does not require sensing that the spectrum is clear before transmitting. \textit{Exponential CSMA} does better still in terms of throughput, although still far off of the performance of the RL agents, but its corresponding network fairness is poor. As noted in Ref. \citenum{Distributed_Fair_Scheduling}, the traditional distributed heuristic algorithms based on CSMA may suffer from unfairness, which we observe here. Five of the \textit{Exponential CSMA} agents capture the channel, as they can all successfully transmit with $k = 5$, and this causes the other five agents' backoff--upper-bound to continually and exponentially increase. The network fairness for the RL agents and \textit{3-persistent}, are both quite good with the best fairness being achieved by the RL agents. The high value for network fairness demonstrates that despite each RL agent being controlled by a distinct model and trying to optimize its own reward, our distributed approach can still lead to cooperative and fair network behavior. 

Additional throughput and fairness benchmarks for 2, 4, and 10 agents with varying values for $k$ can be found in Appendix \ref{Benchmarking Additional Data}. We see that in almost all cases the RL agents outperform the CSMA-inspired benchmarks in terms of throughput. The only time this is not the case is for 10 agents, $k = 1$, $|\mathbb{A}| = 11$, where \textit{exponential CSMA} has higher throughput but much worse fairness. Overall, we take this as confirmation that our heuristic for choosing the size of the action space works well to optimize throughput and fairness separately and to provide a reasonable trade-off when necessary.

\subsection{Buffer Experiments} \label{Buffer Experiments}

In the previous experiments the agents' buffers were incremented on every step, so each agent would always have packets to transmit. While this was useful to determine how the agents would act cooperatively even when all agents had packets to transmit, in many real-world communications systems the information demands are not as stringent or as constant. In the following sections we test the RL agents in settings with less stringent and heterogeneous buffer rates and analyze the performance in terms of throughput, fairness, and how well the agents are able to manage their buffers. In real-world systems, if a buffer reaches its maximum size then information loss or bottlenecks may occur. We compare against the \textit{p-persistent} benchmarking algorithm as in the previous section it showed higher throughput than \textit{p-persistent CSMA} and better fairness than \textit{exponential CSMA}. We test with 4 agents with a maximum buffer size of 100 units, $k = 1$, and set $|\mathbb{A}| = 5$ according to Eqn.~(\ref{eqn:heursitic}).  

\begin{figure}[t]
    \centering
    \subfloat[Network throughput]{
        {\includegraphics[width=8cm]{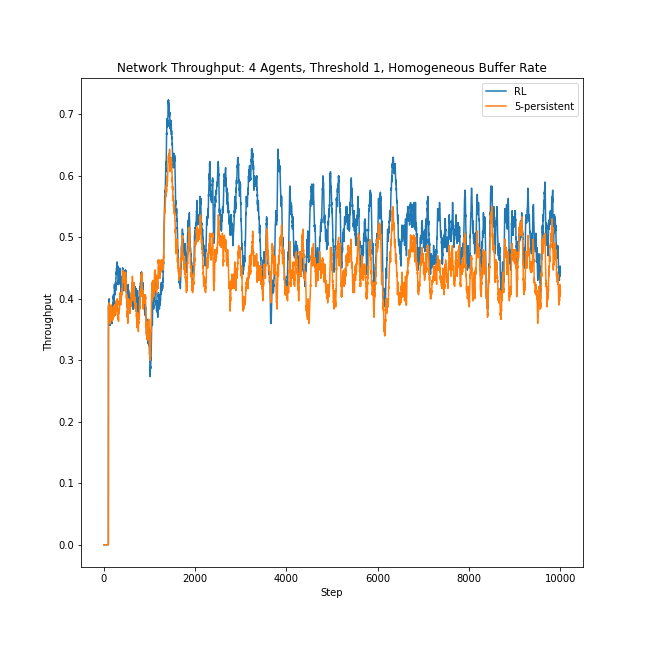}} 
        \label{fig:homogeneous_buffer_throughput}
    }
    \enspace
    \subfloat[Normalized buffer size for Agent 1]{
        {\includegraphics[width=8cm]{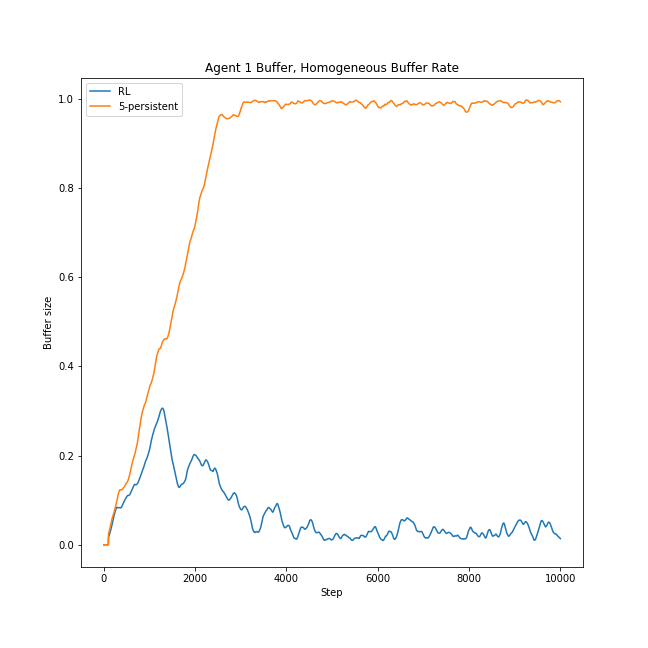}}
        \label{fig:homogeneous_buffer_buffer}
    }
    \caption{Network throughput and normalized buffer size of Agent 1 for a setting where the agents' buffers are homogeneously incremented every 8 steps. There are 4 agents with $k = 1$ and $|\mathbb{A}| = 5$. Network throughput is averaged over 3 runs.}
    \label{fig:homogeneous_buffer}
\end{figure}

In the first experiment we use a homogeneous buffer rate and increment each agent's buffer every 8 steps. Plots for network throughput and the buffer size for one of the agents are displayed in Fig.~\ref{fig:homogeneous_buffer}. The network throughput in Fig.~\ref{fig:homogeneous_buffer_throughput} is clearly, though only slightly higher for the RL agents than for the \textit{5-persistent algorithm}. Despite there not being a large gap in throughput, the RL agents do far better at managing their buffer, as can be seen in Fig.~\ref{fig:homogeneous_buffer_buffer}: both buffers rise initially, but as the training progress the RL agent learns to transmit effectively with the other agents so that the buffer stays low, while the buffer gets maxed-out with the \textit{5-persistent} algorithm. 

Next, we devise an experiment where agents have their buffers incremented heterogeneously. We increment the buffers every $(2, 5, 8, 10)$ steps respectively for each agent. (E.g., agent 1's buffer is incremented every 2 steps, agents 2's buffer is incremented every 5 steps, and so on.) In Fig. \ref{fig:heterogeneous_buffers} in Appendix \ref{Buffer_Experiments_Additional_Data}, the normalized buffer sizes for each of the agents are plotted. For both the RL and \textit{5-persistent} approaches, Agent 1 is not able to prevent its buffer from being maxed-out: the buffer rate is too demanding. For Agent 2, both approaches lead to the buffer being maxed-out initially, but about halfway through the simulation, the RL agent learns to transmit effectively enough so that its buffer stays under maximum buffer size. The buffer sizes for Agent 3 are similar to those in the homogeneous buffer experiments, which makes sense as the buffer rate (incremented every 8 steps) is the same: the RL agent's buffer rises initially, but it quickly learns to act such that the buffer is kept low while for the \textit{5-persistent} algorithm the buffer quickly maxes out. For agent 4, the RL agent handles this buffer rate with ease, and while the \textit{5-persistent} approach manages to slow the increase of the buffer size, the buffer size is still maxed-out at the end (or is at least very close to being maxed-out).

These experiments show our RL approach and heuristic for selecting an action space continues works well even when applied to settings with different, and potentially heterogeneous, information demands.

Additional plots for throughput and fairness for both buffer experiments are shown in Appendix \ref{Buffer_Experiments_Additional_Data}.

\section{Conclusions and Future Directions}
\subsection{Summary}
In this work we presented a distributed MARL approach for transmission control (an MARL-based MAC multiple access method): we formulated the RL agents, theoretically and empirically evaluated the action space to come up with a heuristic for selecting the size of the action space, benchmarked against CSMA-style algorithms, and tested performance with different buffer rates. Overall, the RL agents outperform the benchmarks in terms of throughput, fairness, and the ability to manage their buffer. We share the open source code for the custom Gym environment, RL agents, and the algorithms used for benchmarking through the link in Appendix~\ref{Link to Software}.

\subsection{Discussion}
The results of the action space analysis in Sec.~\ref{Action Space Analysis} show that sometimes the same reinforcement learning problem can be formulated with different action spaces that are functionally equivalent (meaning that any trajectory of actions that is possible using one action space is also possible in the other), and yet some actions spaces will be better at maximizing an objective. Equation~(\ref{eqn:heursitic}) provides a heuristic for choosing the size of the action space given the structure for the action space formulated in Sec.~\ref{Define Action Space}. Choosing the size of the action space in this way can be viewed as tailoring the action space to the problem within a general structure. Selecting a good action space creates a bias towards selecting actions that are likely to create successful patterns of transmission at a network-level. Our method of theoretically analyzing the action space with the assumption of randomly selected actions combined with empirical testing could be used in other cases where there is freedom in choosing an action space.

The benchmarking in Sec.~\ref{Benchmarking} showed that the RL agents outperform the CSMA-style algorithms, and that this performance gap is consistent for varying number of agents and different threshold values for $k$. The fact that the benchmarking algorithms did not perform well indicates some of the shortcomings of traditional distributed algorithms for transmission control, namely, mediocre throughput performance, or good throughput performance but at the expense of fairness. Meanwhile, the RL agents performed well not only in terms of network throughput, but also in fairness demonstrates the power of a distributed MARL approach, where even though the agents act individually to maximize their reward, cooperative behavior can emerge that maximizes network metrics as well if the MARL approach is constructed properly. 

The advantage provided by the RL agents was demonstrated further in the buffer experiments in Sec. \ref{Buffer Experiments}. In these settings the RL agents were often able to keep there buffer sizes low and prevent the their buffers from reaching the maximum size. In real-world systems if an agents buffer were to be maxed-out that could represent information loss or a major bottleneck in the network, so the ability to keep buffer sizes low confers significant value. These experiments also provide evidence that our action space with the heuristic defined in Eqn. (\ref{eqn:heursitic}) to set the size of the action space works well in different settings. The initial action space analysis in Sec.~\ref{Action Space Analysis} was for a setting where each agent's buffer was incremented on every step. This represents an impossibly high information demand if $k$ is smaller than the number of agents, but in basing our action space analysis off of this setting we find the best solution for the most demanding scenario, that is, when all agents always have a constant stream of new packets to transmit. The tests in Sec. \ref{Buffer Experiments} confirm that the RL approach combined and heuristic continues to perform well even if the information demands are reduced or are heterogeneous.

\subsection{Future Directions}
In this work we dealt with the problem of transmission control/medium access at a highly abstracted level. The potential advantage of this is that our approach may be widely applicably to a variety of communications systems, or potentially of independent interest to those interested in MARL. That being said, if one wanted to deploy our approach on physical communications systems, then one may want a more particularized environment to test the approach under the confines of a specific real-world system. Fortunately, there are well-developed network simulators available. Internally we have tested our approach with ns3-gym \cite{ns3gym} and observed similar performance.

While our approach seems to work well, there may also be room for improvement in the formulation of the RL problem in terms of defining an optimal feature space, action space, and reward function. The buffer experiments in Sec.~\ref{Buffer Experiments} showed good results for the RL agents, however, the reward function used, Eqn.~(\ref{eqn:reward}), is quite simplistic and does not directly use any buffer terms. If one wanted to incentivize certain buffer management (e.g., penalize when the buffer is maxed-out or have the reward be proportional to buffer size), then buffer terms could be added to the reward function accordingly.

We tested with 2, 4, and 10 agents, but it would be interesting to see how the approach scales to larger number of agents and with different threshold values for $k$. Additionally, there may be other algorithms to benchmark against that could perform better than the CSMA-style algorithms we employed. Lastly, further study could be directed towards the exploration of different RL algorithms instead of DQN and/or different hyperparameters for the neural networks.

To make it easier for those interested in further investigating or expanding on this work, we provide the open source software for the Gym environment and RL agents in Appendix~\ref{Link to Software}.

\appendix    

\section{Link to Software} \label{Link to Software}
The custom Gym environment as well as the RL model and benchmarking algorithms can be found at \url{https://github.com/Farquhar13/RL_Transmission_Control}.

\section{DQN Hyperparameters} \label{DQN Hyperparameters}
Each agents is equipped with a separate DQN model, but the hyperparameters are the same for all the models. The DQN model uses a small feed-forward neural network for Q-value estimation. We use an $\epsilon$-greedy exploration strategy where a random action is taken with probability $\epsilon$, and on each step $\epsilon$ is decreased by a factor of $\epsilon_{\text{decay}}$ until $\epsilon \leq \epsilon_\text{min}$. The agent trains on every step. The training batch is constructed by randomly sampling past transitions.

\begin{table}[H]
    \centering
    \begin{tabular}{ |p{6cm}|p{3cm}|}
        \hline
        \multicolumn{2}{|c|}{DQN Hyperparameters} \\
        \hline \hline
        Hyperparameter & Value \\
        \hline
        Input dimension & $|s_t| = 4$ \\
        Hidden layer 1 dimension & 128 \\
        Hidden layer 2 dimension & 256 \\
        Activation function to hidden layers & ReLu \\
        Activation function to output layer & Linear \\
        Starting Epsilon & $\epsilon = 1$ \\
        Epsilon decay & $\epsilon_{\text{decay}} = 0.996$ \\
        Minimum epsilon & $\epsilon_{\text{min}} = 0.05$ \\
        Gamma & $\gamma = 0.99$ \\
        Learning rate & $\alpha = 10^{-4}$ \\
        Batch size & 64 \\
        Loss function & Mean Square Error \\
        Optimizer & Adam\\
        \hline
    \end{tabular} 
    \label{tab:DQN Hyperparameters}
\end{table}

\section{Miscoordination Penalty Probabilities} \label{Miscoordination Penalty Probabilitiess}
A miscoordination will occur when more than $k$ agents transmit at the same time. The probability of miscoordination for 10 agents selecting actions uniformly at random with, $k = 2$, $|\mathbb{A}| = 2$, is calculated in Eqn.~(\ref{eqd:P_miscoordination}) using the probability mass function for the binomial distribution:
\begin{equation}
    P(X = x) = \binom{n}{x} p^x (1 - p)^{n - x}.
\end{equation}
We can calculate the miscoordination penalty when $|\mathbb{A}| = 2$ by letting $n$ be number of agents and $p = 0.5$ be the probability of choosing to transmit. Then,
\begin{equation}
    P(\text{miscoordination}) = \sum_{i = k + 1}^{n} \binom{n}{i} p^i (1 - p)^{n - i} = \sum_{i = k + 1}^{n} \binom{n}{i} 0.5^{n}.
\end{equation}

Calculating the probability of miscoordination for larger action spaces is more complicated, as for $|\mathbb{A}| > 2$ the action space will have some multi-step actions, which means that agents will be taking actions asynchronously. 

We can calculate the probability of miscoordination in different way by taking taking a certain view of the actions. If we map the actions such a $1$ denotes a transmission and a $0$ denotes no transmission, then we can view the actions as lists. The action space for $|\mathbb{A}_m| = m$ can be written:
\begin{gather*}
    a_0 = (0) \\
    a_1 = (1) \\
    a_2 = (0, 1) \\
    a_3 = (0, 0, 1) \\
    \ldots \\
    a_{m - 1} = (\underbrace{0, 0, \ldots, 0}_\text{m - 2 times}, 1).
\end{gather*}
So $\mathbb{A}_m = \{a_i\}_{i = 0}^{m - 1}$. If we are at least $m$ steps into the simulation and assume, as before, that actions are taken uniformly at random, then we can calculate the probability of transmitting on an arbitrary step. Let $c_0$ and $c_1$ respectively be the count of the number of zeros and ones in the lists in $\mathbb{A}$. Then the transmission probability, $p_m$ for $|\mathbb{A}_m| = m$ is,
\begin{equation}
    \text{transmission probability for } \mathbb{A}_m = \frac{c_1}{c_0} = p_m,
\end{equation}
and we find  for $m \geq 2$, $c_0 = 1 + \sum_{j = 2}^{m - 1} (j - 1) =  1 + \sum_{j = 1}^{m - 2} j$, and $c_1 = m - 1$ since there is one transmission for every $a_i$ except for $a_0$. The transmission probability will decrease as the size of the action space $m$ gets larger. Formally, this can be seen by noting that $\sum_{j = 1}^{m - 2} j$ is an arithmetic series and it can be shown that $c_0 = 1 + \frac{m^2 - 3m + 2}{2}$, and thus scales quadratically with the size of the action space $m$, whereas $c_1$ only scales linearly with $m$. Using the fact that actions are taken independently, we now can calculate the probability of miscoordination on an arbitrary step for $n$ agents with threshold $k$ and action size $|\mathbb{A}| = m$ as,

\begin{equation}
    P(\text{miscoordination for } \mathbb{A}_m) = \sum_{i = k + 1}^{n} p_m^i. 
\end{equation}

Assuming that actions are selected uniformly at random, we can now prove that a larger action space leads to a lower miscoordination probability for $n$ agents with threshold $k$. Let $|\mathbb{A}_x| = x$, $|\mathbb{A}_x| = y$ where $x < y$. We have shown that a larger action space leads to a smaller transmission probability, so $p_x > p_y$. Suppose $P(\text{miscoordination for } \mathbb{A}_x) > P(\text{miscoordination for } \mathbb{A}_y)$. So,
\begin{equation}
\begin{aligned}
P(\text{miscoordination for } \mathbb{A}_x) &> P(\text{miscoordination for } \mathbb{A}_y) \\
\sum_{i = k + 1}^{n} p_x^i &> \sum_{i = k + 1}^{n} p_y^i \\
\sum_{i = k + 1}^{n} p_x^i - \sum_{i = k + 1}^{n} p_y^i &> 0\\
\sum_{i = k + 1}^{n} (p_x^i - p_y^i) &> 0.
\end{aligned}
\end{equation}

The last line is true since, for any $a, b, c \in \mathbb{R}^+$ with $a > b$, then $a^c > b^c$, so $a^c - b^c > 0$, thus every $(p_x^i - p_y^i)$ is positive and a sum of positive terms is greater than zero. Therefore, $P(\text{miscoordination for } \mathbb{A}_x) > P(\text{miscoordination for } \mathbb{A}_y)$. $\blacksquare$ 

Put simply, a larger action space will result in a lower transmission probability, which implies a lower probability for miscoordination.

\section{Action Space Analysis Additional Data} \label{Action Space Analysis Additional Data}

\begin{table}[H]
    \centering
    \begin{tabular}{ |p{3cm}||p{3cm}|p{3cm}|p{3cm}|  }
        \hline
        \multicolumn{4}{|c|}{Throughput, $k = 1$} \\
        \hline \hline
        No. of Agents / No. of Actions& 2 Agents & 4 Agents & 10 Agents\\
        \hline
        2 Actions & Mean = 0.78006   STDEV = 0.17884 & Mean = 0.60261   STDEV = 0.03866 & Mean = 0.57710     STDEV = 0.03215 \\
        \hline
        3 Actions & Mean = 0.95078   STDEV = 0.02402 & Mean = 0.61586   STDEV = 0.02624 & Mean = 0.46621     STDEV = 0.06823 \\
        \hline
        4 Actions & Mean = 0.64971   STDEV = 0.01450 & Mean = 0.63784   STDEV = 0.03712 & Mean = 0.48662   STDEV = 0.06416 \\
        \hline
        5 Actions & Mean = 0.73348   STDEV = 0.06607 & Mean = 0.69083   STDEV = 0.17607 & Mean = 0.44202     STDEV = 0.03817 \\
        \hline
        6 Actions & Mean = 0.81372   STDEV = 0.07375 & Mean = 0.66441   STDEV = 0.01132 & Mean = 0.40869   STDEV = 0.05880 \\
        \hline
        7 Actions & \center -- & Mean = 0.67896   STDEV = 0.12725 & Mean = 0.43054         STDEV = 0.05287 \\
        \hline
        8 Actions  & \center -- & Mean = 0.69733   STDEV = 0.07090 & Mean = 0.40404            STDEV = 0.05060 \\
        \hline
        9 Actions & \center -- & \center -- & Mean = 0.37100  STDEV = 0.02976 \\
        \hline
        10 Actions & \center -- & \center -- & Mean = 0.41004    STDEV = 0.07357 \\
        \hline
        11 Actions & \center -- & \center -- & Mean = 0.43033    STDEV = 0.05623\\
 
        \hline
    \end{tabular} 
    \caption{Average \& standard deviation of throughput for threshold 1. Data is from the last 1,000 steps of a 10,000 step training period and averaged over three runs.}
    \label{tab:throughput_threshold1}
\end{table}

\begin{table}[H]
    \centering
    \begin{tabular}{ |p{3cm}||p{3cm}|p{3cm}|p{3cm}|  }
        \hline
        \multicolumn{4}{|c|}{Fairness, $k = 1$} \\
        \hline \hline
        No. of Agents / No. of Actions& 2 Agents & 4 Agents & 10 Agents\\
        \hline
        2 Actions & Mean = 0.87845   STDEV = 0.20949 & Mean = 0.52672   STDEV = 0.03866 & Mean = 0.29744   STDEV = 0.04632 \\
        \hline
        3 Actions & Mean = 0.99983   STDEV = 0.00010 & Mean = 0.67614   STDEV = 0.13102 & Mean = 0.42953         STDEV = 0.03979 \\
        \hline
        4 Actions & Mean = 0.99976   STDEV = 0.00001 & Mean = 0.83247   STDEV = 0.00112 & Mean = 0.49201            STDEV = 0.02939\\
        \hline
        5 Actions & Mean = 0.99159   STDEV = 0.00820 & Mean = 0.93710   STDEV = 0.06327 & Mean = 0.55159            STDEV = 0.07052 \\
        \hline
        6 Actions & Mean = 0.99634   STDEV = 0.00291 & Mean = 0.88923   STDEV = 0.06939 & Mean = 0.63227            STDEV = 0.02151 \\
        \hline
        7 Actions & \center -- & Mean = 0.90435   STDEV = 0.10433 & Mean = 0.65110            STDEV = 0.05479  \\
        \hline
        8 Actions  & \center -- & Mean = 0.97265   STDEV = 0.00634 & Mean = 0.67006            STDEV = 0.02951 \\
        \hline
        9 Actions & \center -- & \center -- & Mean = 0.67199            STDEV = 0.04562 \\
        \hline
        10 Actions & \center -- & \center -- & Mean = 0.69593            STDEV = 0.02341 \\
        \hline
        11 Actions & \center -- & \center -- & Mean = 0.67655            STDEV = 0.07446 \\
 
        \hline
    \end{tabular} 
    \caption{Average \& standard deviation of fairness for threshold 1. Data is from the last 1,000 steps of a 10,000 step training period and averaged over three runs.}
    \label{tab:fairness_threshold1}
\end{table}

\begin{table}[H]
    \centering
    \begin{tabular}{ |p{3cm}||p{3cm}|p{3cm}|}
        \hline
        \multicolumn{3}{|c|}{4 Agents, Throughput \& Fairness, $k = 3$} \\
        \hline \hline
        No. of Action & Throughput & Fairness\\
        \hline
        2 Actions & Mean = 1.71945   STDEV = 0.15431 & Mean = 0.89216   STDEV = 0.03130 \\
        \hline
        3 Actions & Mean = 1.99699   STDEV = 0.02159 & Mean = 0.99979   STDEV = 0.00009 \\
        \hline
        4 Actions & Mean = 1.61192   STDEV = 0.04602 & Mean = 0.99045   STDEV = 0.00238 \\
        \hline
        5 Actions & Mean = 1.55774   STDEV = 0.06220 & Mean = 0.98197   STDEV = 0.00516 \\
        \hline
        6 Actions & Mean = 1.57845   STDEV = 0.08176 & Mean = 0.94473   STDEV = 0.02596 \\
        \hline
        7 Actions & Mean = 1.45513   STDEV = 0.08087 & Mean = 0.94238   STDEV = 0.00399 \\
        \hline
        8 Actions & Mean = 1.56845   STDEV = 0.08251 & Mean = 0.90685   STDEV = 0.02018 \\

        \hline
    \end{tabular} 
    \caption{Average \& standard deviation of throughput \& fairness for 4 agents threshold 3. Data is from the last 1,000 steps of a 10,000 step training period and averaged over three runs.}
    \label{tab:throughput_fairness_4agent_threshold3}
\end{table}

\begin{table}[H]
    \centering
    \begin{tabular}{ |p{3cm}||p{3cm}|p{3cm}|}
        \hline
        \multicolumn{3}{|c|}{10 Agents, Throughput \& Fairness, $k = 5$} \\
        \hline \hline
        No. of Action & Throughput & Fairness\\
        \hline
        2 Actions & Mean = 3.19046   STDEV = 0.58891 & Mean = 0.85694   STDEV = 0.14962 \\
        \hline
        3 Actions & Mean = 3.91997   STDEV = 0.03990 & Mean = 0.98970   STDEV = 0.00431 \\
        \hline
        4 Actions & Mean = 3.31367   STDEV = 0.00995 & Mean = 0.99861   STDEV = 0.00126 \\
        \hline
        5 Actions & Mean = 2.59365   STDEV = 0.17936 & Mean = 0.98708   STDEV = 0.00375 \\
        \hline
        6 Actions & Mean = 2.43987   STDEV = 0.08961 & Mean = 0.96973   STDEV = 0.01043 \\
        \hline
        7 Actions & Mean = 2.41671   STDEV = 0.15506 & Mean = 0.96493   STDEV = 0.00990 \\
        \hline
        8 Actions & Mean = 2.44283   STDEV = 0.16497 & Mean = 0.92002   STDEV = 0.00580 \\
        \hline
        9 Actions & Mean = 1.85070   STDEV = 0.05589 & Mean = 0.94923   STDEV = 0.01256 \\
        \hline
        10 Actions & Mean = 1.85649   STDEV = 0.07320 & Mean = 0.90869   STDEV = 0.02141 \\
        \hline
        11 Actions & Mean = 1.70585   STDEV = 0.08802 & Mean = 0.92496   STDEV = 0.01213 \\
        \hline
    \end{tabular} 
    \caption{Average \& standard deviation of throughput \& fairness for 10 agents threshold 5. Data is from the last 1,000 steps of a 10,000 step training period and averaged over three runs.}
    \label{tab:throughput_fairness_10agent_threshold5}
\end{table}

\section{Benchmarking Additional Data} \label{Benchmarking Additional Data}
\begin{figure}[H]
    \centering
    \includegraphics[width=.45\textwidth]{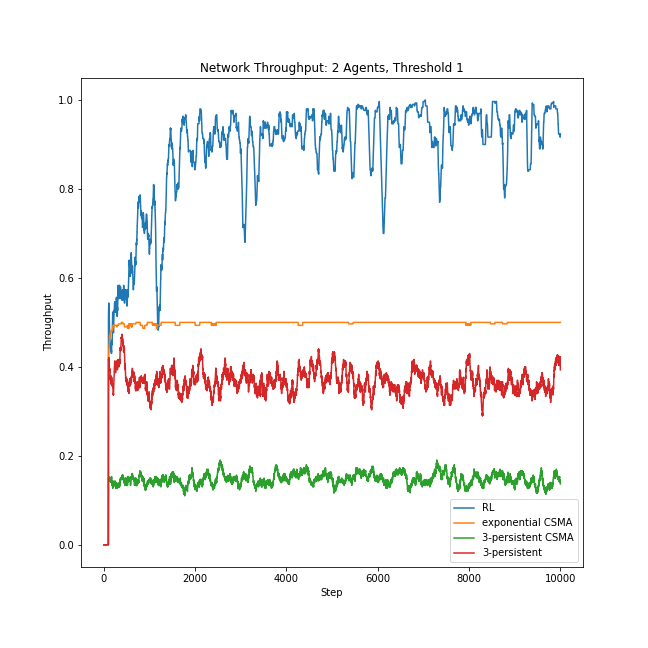}
    \includegraphics[width=.45\textwidth]{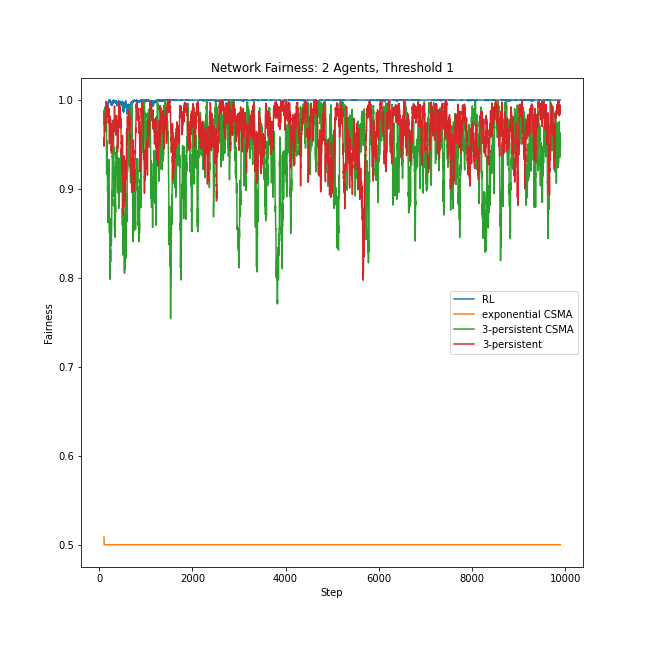}
    \caption{Network throughput and fairness for 2 agents, $k = 1$, $|\mathbb{A}| = 3$. (Data averaged over three runs.)}
\end{figure}

\begin{figure}[H]
    \centering
    \includegraphics[width=.45\textwidth]{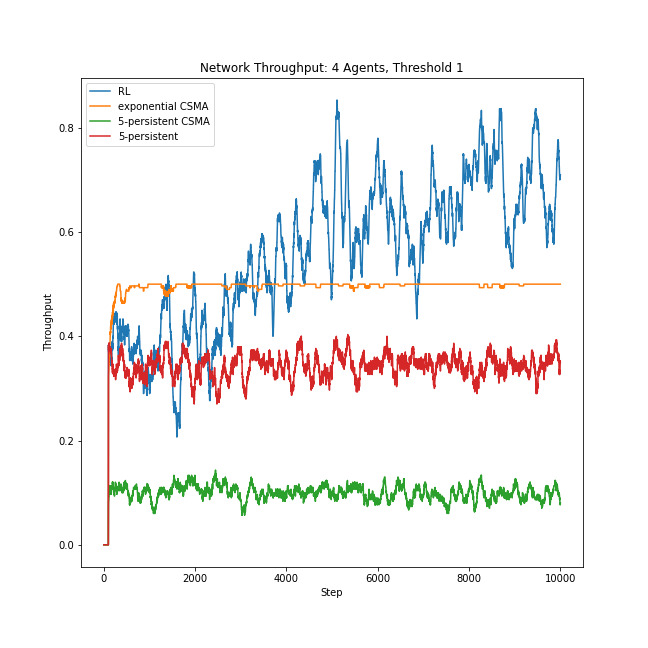}
    \includegraphics[width=.45\textwidth]{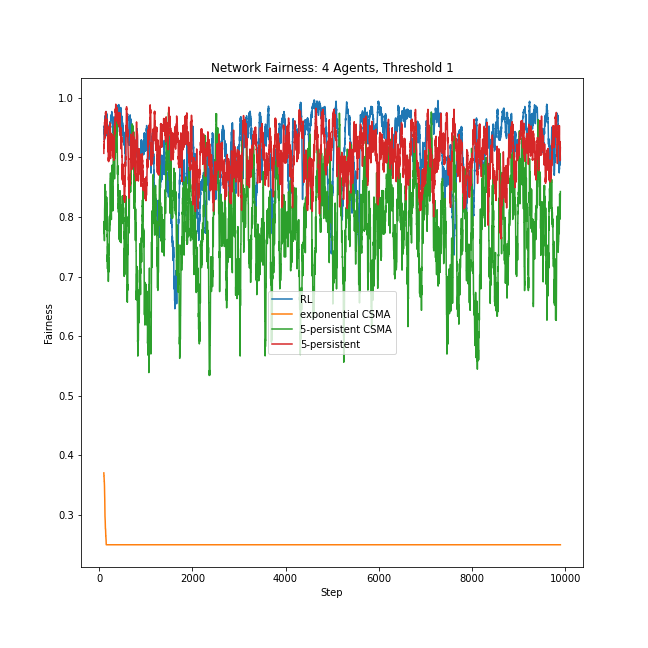}
    \caption{Network throughput and fairness for 4 agents, $k = 1$, $|\mathbb{A}| = 5$. (Data averaged over three runs.)}
\end{figure}

\begin{figure}[H]
    \centering
    \includegraphics[width=.45\textwidth]{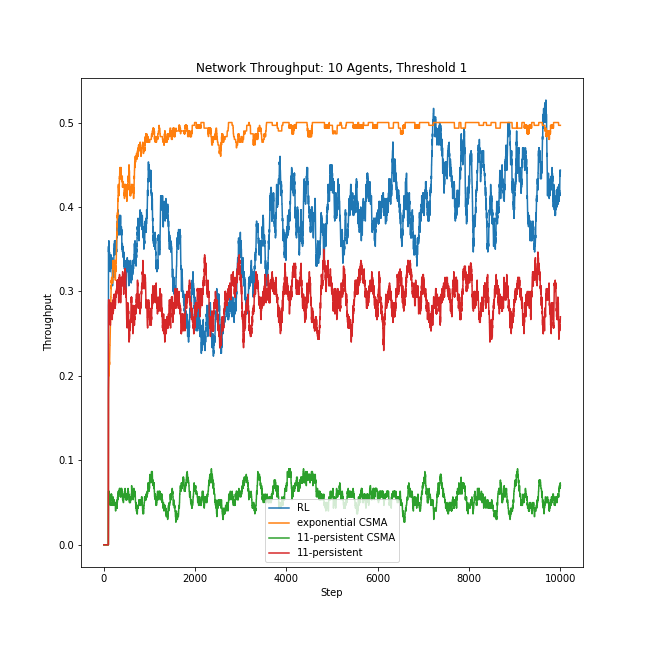}
    \includegraphics[width=.45\textwidth]{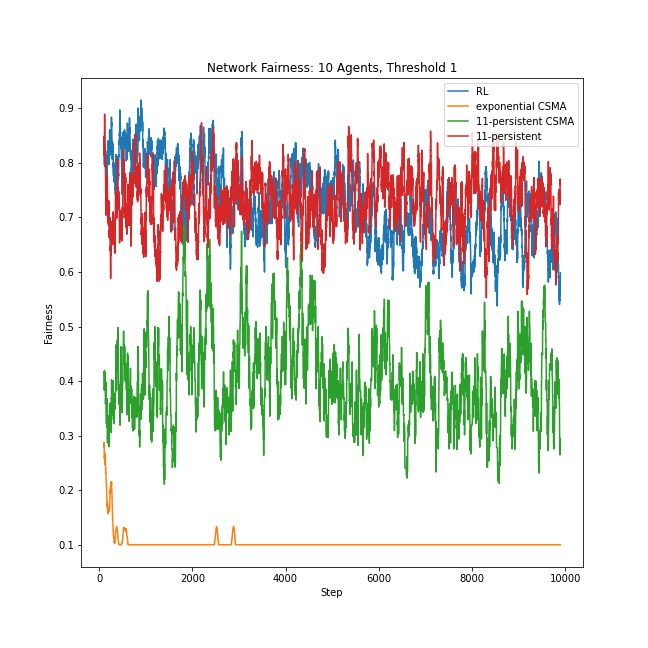}
    \caption{Network throughput and fairness for 10 agents, $k = 1$, $|\mathbb{A}| = 11$. (Data averaged over three runs.)}
\end{figure}

\begin{figure}[H]
    \centering
    \includegraphics[width=.45\textwidth]{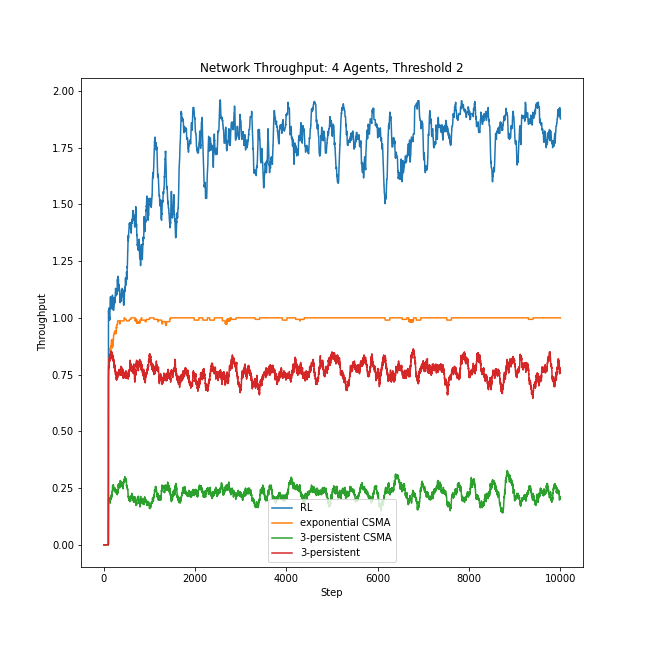}
    \includegraphics[width=.45\textwidth]{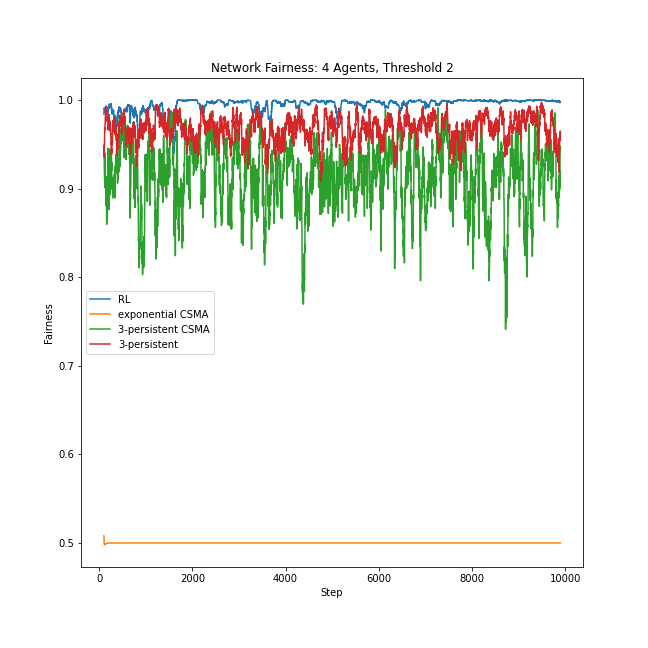}
    \caption{Network throughput and fairness for 4 agents, $k = 2$, $|\mathbb{A}| = 3$. (Data averaged over three runs.)}
\end{figure}

\begin{figure}[H]
    \centering
    \includegraphics[width=.45\textwidth]{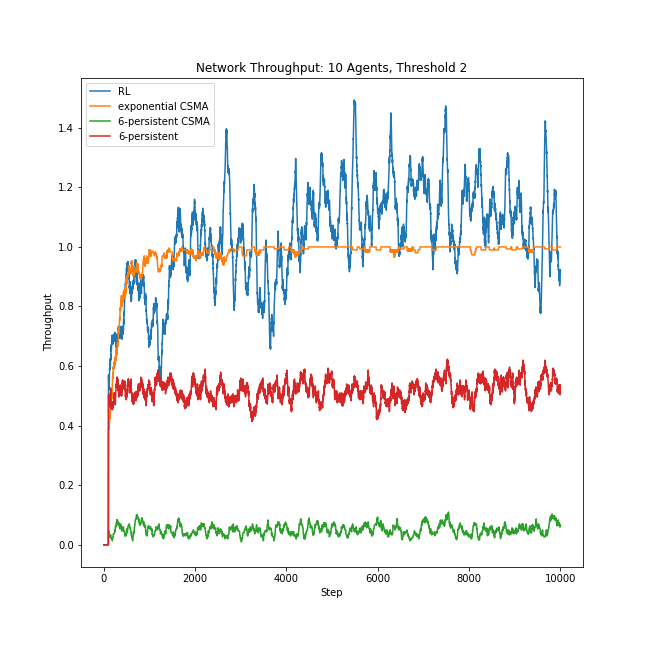}
    \includegraphics[width=.45\textwidth]{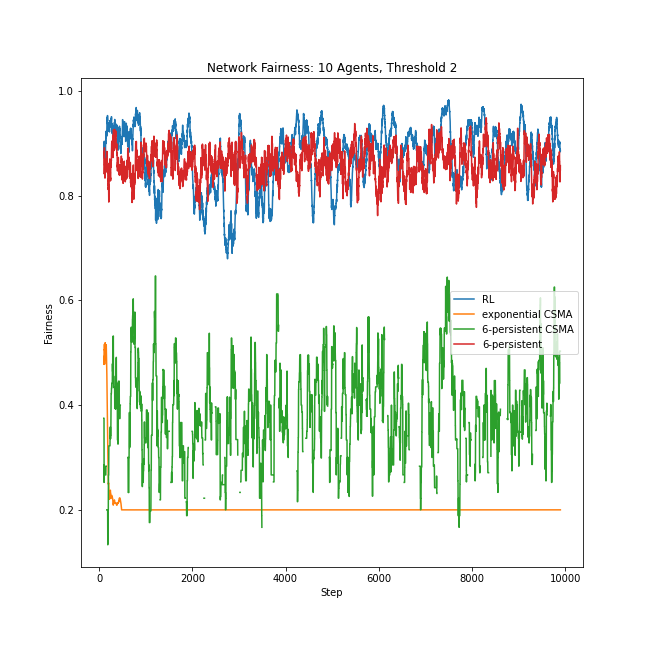}
    \caption{Network throughput and fairness for 10 agents, $k = 2$, $|\mathbb{A}| = 6$. (Data averaged over three runs.)}
\end{figure}

\section{Buffer Experiments Additional Data} \label{Buffer_Experiments_Additional_Data}
\begin{figure}[H]
    \centering
    \includegraphics[width=0.7\textwidth]{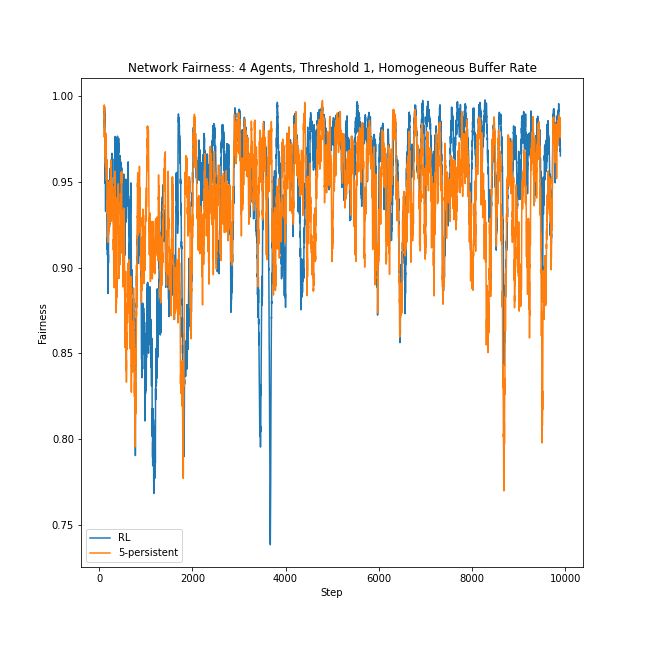}
    \caption{Network fairness for 4 agents, $k = 1$, $|\mathbb{A}| = 5$, where each agent's buffer is incremented every 8 steps. (Fairness averaged over three runs.)}
\end{figure}

\begin{figure}[H]
    \centering
    \includegraphics[width=.45\textwidth]{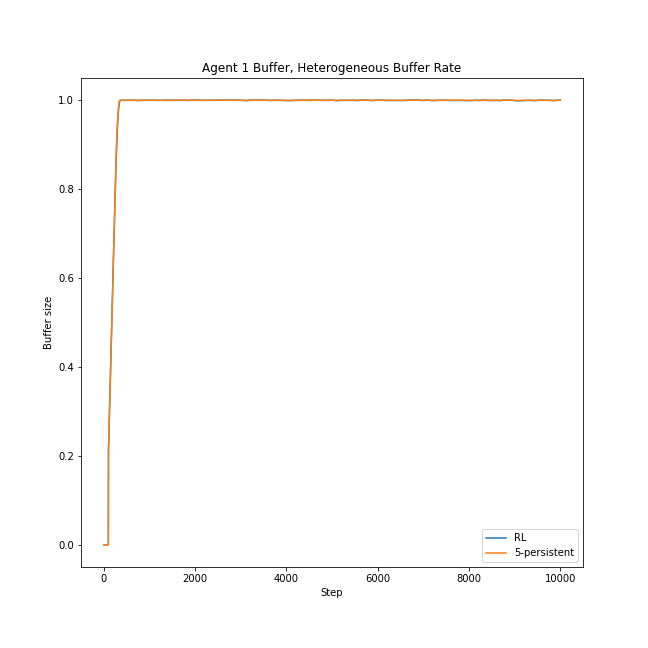}
    \includegraphics[width=.45\textwidth]{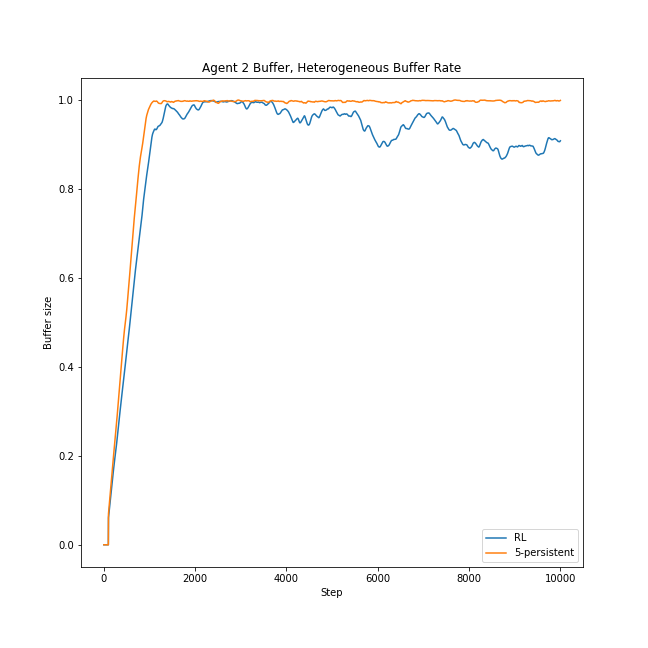}
    \includegraphics[width=.45\textwidth]{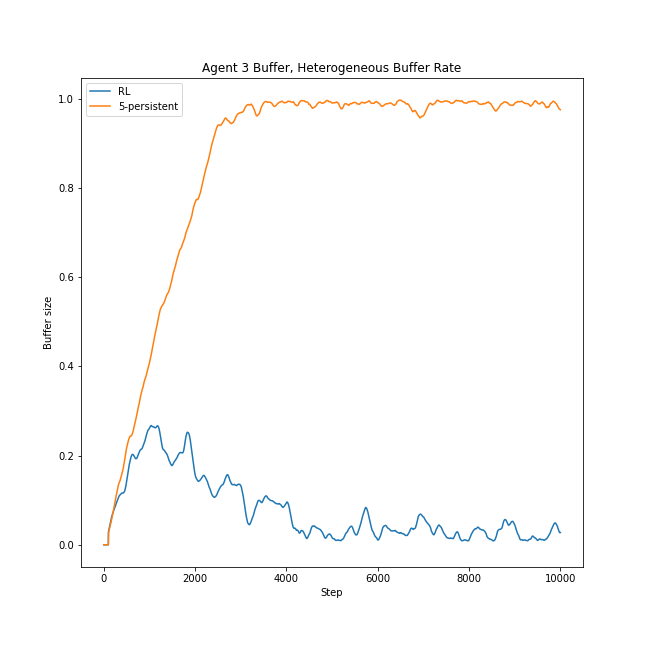}
    \includegraphics[width=.45\textwidth]{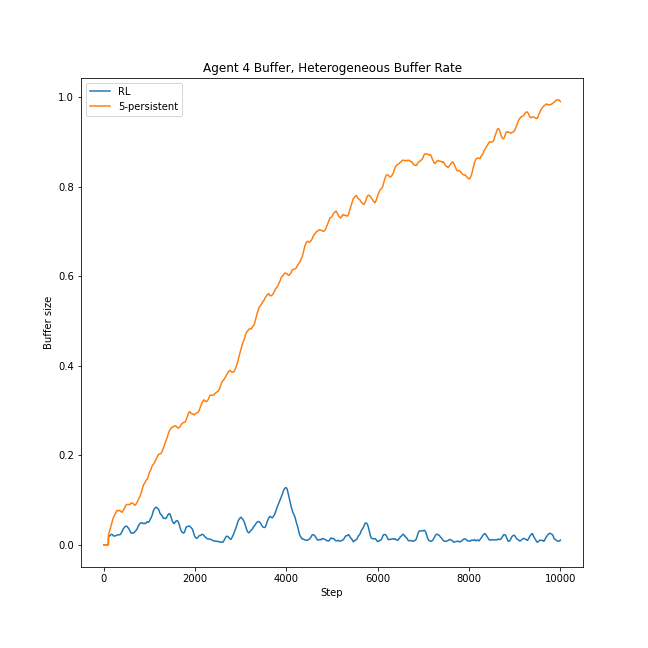}
    \caption{Buffer sizes for 4 agents, k = 1, $|A| = 5$, where buffers are incremented heterogeneously every (2, 5, 8, 10) steps respectively for each agent. (Buffer sizes averaged over three runs.)}
    \label{fig:heterogeneous_buffers}
\end{figure}

\begin{figure}[H]
    \centering
    \includegraphics[width=.45\textwidth]{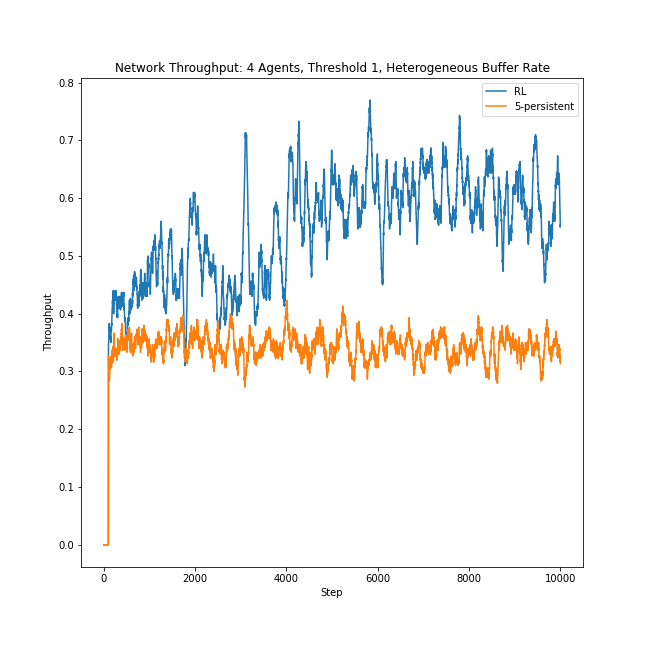}
    \includegraphics[width=.45\textwidth]{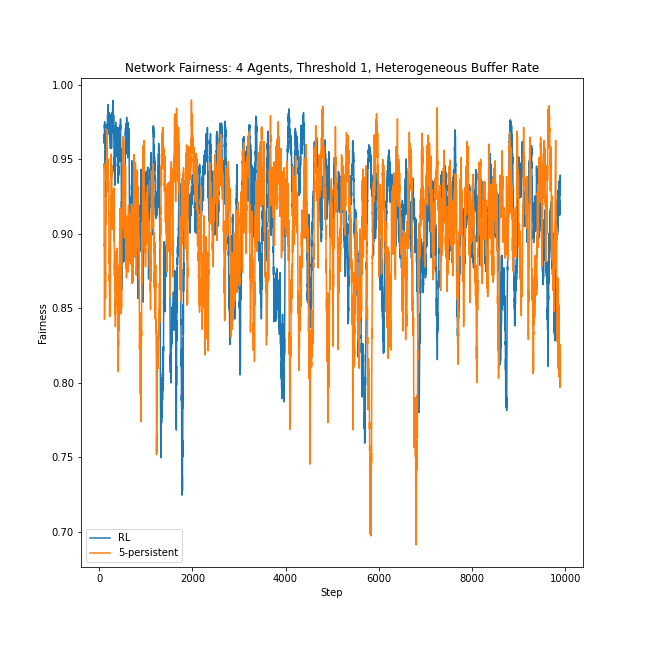}
    \caption{Network throughput and fairness for 4 agents, $k = 1$, $|\mathbb{A}| = 5$, where buffers are incremented heterogeneously every (2, 5, 8, 10) steps respectively for each agent. (Throughput and fairness averaged over three runs.)}
    \label{fig:heterogeneous_buffer_throughput_fairness}
\end{figure}

\bibliography{main} 
\bibliographystyle{spiebib} 

\end{document}